\documentclass[11pt,a4paper]{article}
\usepackage[hyperref]{acl2019}
\usepackage{times}
\usepackage{latexsym}

\usepackage{url}
\usepackage{graphicx}
\usepackage{booktabs}
 \usepackage{todonotes}

\aclfinalcopy 

\title{Fine-tune Bert for DocRED with Two-step Process}

\author{
 Hong Wang$^\dagger$,
 Christfried Focke$^{\ddag}$, 
 Rob Sylvester$^{\ddag}$, 
 Nilesh Mishra$^\ddag$,
 William Wang$^\dagger$
\\
 $^\dagger$ University of California, Santa Barbara\\
 $^\ddag$ LogMeIn\\
 \{hongwang600, william\}@cs.ucsb.edu,\\
 \{Christfried.Focke, Rob.Sylvester, Nilesh.Mishra\}@logmein.com
 }

\date{}

\begin{document}
\maketitle
\begin{abstract}
  Modelling relations between multiple entities has attracted increasing attention recently, and a new dataset called DocRED has been collected in order to accelerate the research on the document-level relation extraction. Current baselines for this task uses BiLSTM to encode the whole document and are trained from scratch. We argue that such simple baselines are not strong enough to model to complex interaction between entities. In this paper, we further apply a pre-trained language model (BERT) to provide a stronger baseline for this task. We also find that solving this task in phases can further improve the performance. The first step is to predict whether or not two entities have a relation, the second step is to predict the specific relation \footnote{Code can be found in \url{https://github.com/hongwang600/DocRed}}.
\end{abstract}

\section{Introduction}

The task of relation extraction aims to automatically identify relationships between entities. It has been proven to be essential for many downstream applications such as question answering \cite{KBQA_Yih,KBQA_Mo}. Previous research \cite{SocherHMN12,ZengLLZZ14,ZengLC015,SantosXZ15,XiaoL16,BiLSTM_RE,LinSLLS16,WuBR17,WangXQ18a,HanYLSL18,mo_acl19} on relation extraction mainly focuses on sentence-level, i.e., predicting the relation for entities in a given sentence. Recently the large-scale document-level relation extraction dataset DocRED \cite{yao2019DocRED} was published, which requires the model to predict a relation for every pair of entities in a document. This setting is more challenging, since a large number of relational facts are expressed across multiple sentences, and modeling of complex interactions between entities is required.

In \cite{yao2019DocRED} several baselines for the document-level Relation Extraction (RE) task are presented. The best model uses a BiLSTM \cite{BiLSTM} to encode the whole document, entities are represented by their average word embedding. A BiLinear layer is then applied to predict the relation for a given entity pair. However, we argue that a pre-trained language model, such as BERT \cite{BERT}, can provide a further boost in performance, since it already captures important language features and may capture some common sense knowledge.

In this paper, we use BERT to encode the document. A BiLinear layer is applied to predict the relation between entity pairs. We fine-tune the whole model using annotated data in the DocRED dataset, which increases the F1 score by about $2\%$. We also found that modeling the document-level relation extraction through a two-step process can further improve the performance. The first step is to predict whether a pair of entities has a relation or not. The second step is to predict the specific relation for a given entity pair. Note that the model we use in the second step is trained with pairs that have relations annotated in DocRED.

\section{Model}
In this section, we will first introduce the BERT model for document-level RE, then we will explain how to use the two-steps training process to further improve the performance.

\subsection{BERT Model}
\begin{figure}
    \centering
    \includegraphics[width=0.45\textwidth]{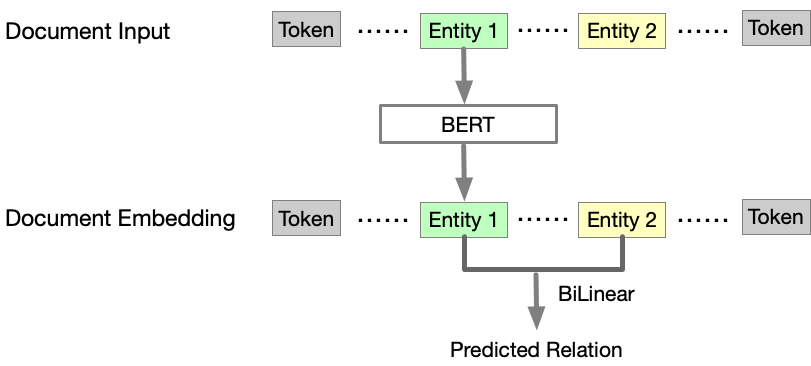}
    \caption{BERT model for DocRED.}
    \label{fig:model}
\end{figure}

Let $[x_1, x_2,\cdots, x_n]$ denote the document input, and $[e_1,e_2,\cdots,e_m]$ denote the $m$ entities in the document. We use BERT to encode the document as follows:
$$
    [h_1, h_2, \cdots, h_n] = \textsc{BERT}([x_1,x_2,\cdots,x_n]),
$$
from which we can get the embeddings $[h_{e_1}, h_{e_2},\cdots,h_{e_m}]$. Then for each pair of entities $(e_i, e_j)$, we use BiLinear layer to predict its relation:
$$
    r_{i,j} = \textsc{BiLinear}(h_{e_i}, h_{e_j}).
$$
The whole model structure is represented in Figure \ref{fig:model}. We use BERT-base in our experiments. 

\subsection{Two-step Training Process}
In the DocRED dataset, there is no relation for most entity pairs, which causes a large label imbalance, i.e., most entity pairs belong to the N/A relation. To alleviate this problem, we use a two-step training process.

In the first step, we only identify whether or not there exists a relation between a given entity pair, i.e., we simplify the problem to a binary classification problem. We use BERT for this step as mentioned above,  where all of the annotated data is used to train the model. Sub-sampling is applied to balance relational and N/A pairs in each batch.

In the second step, we learn a model to identify the specific relationship between a given pair of entities. The model structure is the same BERT model as in the first step. The difference lies in the training data and labels: We only use these relational facts (i.e., entity pairs with relations) to train the model, so that the model can learn to distinguish between these different relations. Empirically we found that the second step is relatively easy, as we achieve about $90\%$ accuracy. The bottleneck of the problem lies in the first step, which is to distinguish whether there is a relation or not.

After the two-step training, the testing process is straight forward. For a given pair of entities, the model from the first step is first applied to predict whether there is a relation between them. If it predicts a relation, then the model from the second step is applied to predict a specific relation.

\section{Experiments}
In this section, we will introduce the DocRED dataset, our implementation details of the BERT model, and the experimental results.

\subsection{DocRED Dataset}
The DocRED dataset is collected through distant supervision \cite{DistantSupervision} on Wikipedia documents and Wikidata. The named entity recognition is performed first on each document. Then the identified entities are linked to Wikidata items, and entities with same Knowledge Base (KB) ID are merged. Finally, the relations for entity pairs are obtained by querying Wikidata. Further processing, like named entity and coreference annotation, entity linking, and relation and supporting evidence collection, is conducted based on the collected distantly supervised data. We refer the readers to the original paper \cite{yao2019DocRED} for more details.

The DocRED dataset has wide coverage over a variety of topics. The entity types include person, location, organization, time, number and miscellaneous entity names. The relation types include science, art, time, personal life, etc. In order to be successful on this dataset, a variety of reasoning types are required, including pattern recognition, logical reasoning, coreference reasoning, and common-sense reasoning. The DocRED dataset provides both annotated training data (sampled from collected distantly supervised and humanly labeled data) and distantly supervised data. In our experiments, we only use the annotated data. Statistics about the dataset are listed in Table \ref{tab:statics}.

\begin{table}[]
    \centering
    \begin{tabular}{c|c|c|c|c}
    \toprule
        Setting & \# Doc & \# Rel & \# Inst & \#Fact \\\midrule
        Train & 3,053 & 96 & 38,269 & 34,715 \\
        Dev & 1,000 & 96 & 12,332 & 11,790 \\
        Test & 1,000 & 96 & 12,842 & 12,101\\\bottomrule
    \end{tabular}
    \caption{Statistics of the dataset. \# Doc,  \# Rel, \# Inst, \#Fact denote the number of documents, the number of relations, the number of relation instances and the number of relational facts respectively.}
    \label{tab:statics}
\end{table}

\subsection{Implementation Details}
We use BERT-base in our experiments. The learning rate is set to $10^{-5}$. The embedding size of BERT model is $768$. A transformation layer is used to project the BERT embedding into a low-dimensional space of size $128$. In the low-dimension space, a BiLinear layer is applied to predict the relation for a given entity pair.

In the first step, we set the relation label for all relational instances to be $1$, while the label for all N/A relations to be $0$. We randomly sample N/A relations at a ratio $3:1$ within a batch. In the second step, we train a new model using only relational instances, and the specific relation label is kept in this step.

\subsection{Results}
We compare the BERT model with several baselines presented in \cite{yao2019DocRED} including a CNN \cite{CNN}, LSTM \cite{BiLSTM}, bidirectional LSTM (BiLSTM) \cite{BiLSTM_RE} and Context-Aware models.
The first three models differ from the BERT model in the encoder, i.e., they use CNN, LSTM, and BiLSTM as encoder respectively. Details about Context-Aware model can be found in \cite{Context_aware}.

The main results are presented in Table \ref{tab:result}. We can see that we obtain a $2\%$ $F_1$ improvement by using the BERT encoder, which indicates that it may contain useful information such as common-sense knowledge in order to solve this task. By using the two-step training process, performance is improved further improve. In our experiments, we find that the accuracy for the second step is above $90\%$, which means the bottleneck lies in the first step, e.g., predicting whether a relation exists for a given entity pair.

\begin{table}[]
    \centering
    \begin{tabular}{c|c|c}\toprule
         Model & Dev & Test  \\\midrule
         CNN & 43.45 & 42.26\\
         LSTM & 50.68 & 50.07 \\
         BiLSTM & 50.94 & 51.06\\
         Context-Aware & 51.09 & 50.70\\\midrule
         BERT & 54.16 & 53.20 \\
         BERT-Two-Step & \bf 54.42 & \bf 53.92 \\\bottomrule
    \end{tabular}
    \caption{Comparison of the BERT model with other baselines. We report $F_1$ score on the Dev and Test set.}
    \label{tab:result}
\end{table}{}

\subsection{Complex interaction modeling}
To test whether current model can capture the complex interaction between entities, we use a SentModel which encodes the document sentence by sentence. Then we locate each entity within a specific sentence and compute its embedding by averaging the word embedding of the entity name. 
In this way, there will be no interaction between sentences since we encode the whole document sentence by sentence. We present the results in Table \ref{tab:sent}. Surprisingly, the SentModel can achieve very similar performance compared to the BiLSTM model which encodes the whole document as a sequence. Therefore, the current model fails to capture complex interactions among entities, and only local information around each entity is used to predict a relation.

\begin{table}[]
    \centering
    \begin{tabular}{c|c|c}\toprule
        Model & F1 & AUC \\\midrule
        BiLSTM & 50.94 & 50.26\\
        SentModel & 50.97 & 49.31 \\\bottomrule
    \end{tabular}
    \caption{Comparison of the BiLSTM baseline with SentModel which encode the document sentence by sentence. We report the F1 score and AUC on the Dev set here.}
    \label{tab:sent}
\end{table}{}

\section{Conclusion \& Discussion}
In this paper, we investigate the usage of BERT for document-level RE. We find that BERT can improve the performance significantly, which we think may benefit from the common sense knowledge learned during pre-training. We also find that using a two-step training process can further improve the performance. The difficulty of this dataset is to distinguish whether there exists a relation between a pair of entities, while identifying a specific relation seems to be less challenging. Another discovery is that current models fail to model complex interaction between entities, which we think is the key to solve the problem of document-level RE.

\bibliography{acl2019}
\bibliographystyle{acl_natbib}

\appendix

\end{document}